\documentclass{article}
\usepackage{amsmath,amsfonts}
\usepackage{algorithmic}
\usepackage{array,multirow}
\usepackage{textcomp}
\usepackage{stfloats}
\usepackage{url}
\usepackage{verbatim}
\usepackage{graphicx}
\usepackage{booktabs}
\usepackage{fancyhdr}
\usepackage{floatrow}
\usepackage[utf8]{inputenc} % allow utf-8 input
\usepackage[T1]{fontenc}    % use 8-bit T1 fonts
\usepackage{hyperref}       % hyperlinks
\usepackage{url}            % simple URL typesetting
\usepackage{booktabs}       % professional-quality tables
\usepackage{amsfonts}       % blackboard math symbols
\usepackage{nicefrac}       % compact symbols for 1/2, etc.
\usepackage{microtype}      % microtypography
\usepackage{xcolor}         % colors
\usepackage{amsmath,amssymb,amsfonts,amsthm}
\usepackage{graphicx,xcolor}
\usepackage{wrapfig}
\usepackage{caption}
\usepackage{subcaption}
\usepackage{floatrow} 
\usepackage{multirow}
\usepackage{array}
\usepackage{cite}
\newcolumntype{P}[1]{>{\centering\arraybackslash}p{#1}}
\usepackage[framemethod=tikz]{mdframed}
\newtheorem{definition}{Definition}
\newtheorem{example}{Example}
\usepackage[a4paper, total={6in, 8in}]{geometry}
\newgeometry{top=1in,bottom=1in,right=1.5in,left=1.5in}

\makeatletter
\newcommand*\titleheader[1]{\gdef\@titleheader{#1}}
\AtBeginDocument{%
  \let\st@red@title\@title
  \def\@title{%
    \bgroup\normalfont\large\centering\@titleheader\par\egroup
    \vskip0.0em\st@red@title}
}
\makeatother

\title{A Canonical Data Transformation for Achieving Inter- and Within-group Fairness}

\titleheader{\small This work has been submitted to the IEEE for possible publication. Copyright may be transferred without notice, after which this version may no longer be accessible.}

\author{Zachary McBride Lazri\thanks{Zachary Lazri, Dana Dachman-Soled, and Min Wu are with the University of Maryland, College Park
(e-mail: zlazri@umd.edu; danadach@umd.edu; minwu@umd.edu).}, \hspace{-7mm} \and
        Ivan Brugere\thanks{Ivan Brugere, Antigoni Polychroniadou, and Danial Dervovic are with JPMorgan A.I. Research
(e-mail: ivan.brugere@jpmchase.com; antigoni.polychroniadou@jpmorgan.com; danial.dervovic@jpmchase.com).}, \hspace{-7mm} \and
        Xin Tian \thanks{Xin Tian was with the 
University of Maryland, College Park when this work was started and is now with Meta, Menlo Park, CA 94025 USA (e-mail: xtian17@terpmail.umd.edu)}, \hspace{-7mm} \and Dana Dachman-Soled\footnotemark[1], \and Antigoni Polychroniadou\footnotemark[2],  \hspace{-10mm} \and Danial Dervovic\footnotemark[2],  \hspace{-10mm} \and 
        and~Min~Wu\footnotemark[1]
}
\date{}

\begin{document}

\maketitle

\begin{abstract}
Increases in the deployment of machine learning algorithms for applications that deal with sensitive data have brought attention to the issue of fairness in machine learning. Many works have been devoted to applications that require different demographic groups to be treated fairly. However, algorithms that aim to satisfy inter-group fairness (also called group fairness) may inadvertently treat individuals within the same demographic group unfairly. To address this issue, this paper introduces a formal definition of \textit{within-group fairness} that maintains fairness among individuals from within the same group. A pre-processing framework is proposed to meet both inter- and within-group fairness criteria with little compromise in performance. The framework maps the feature vectors of members from different groups to an inter-group fair canonical domain before feeding them into a scoring function. The mapping is constructed to preserve the relative relationship between the scores obtained from the unprocessed feature vectors of individuals from the same demographic group, guaranteeing within-group fairness. This framework has been applied to the Adult, COMPAS risk assessment, and Law School datasets, and its performance is demonstrated and compared with two regularization-based methods in achieving inter-group and within-group fairness.
\end{abstract}

\section{Introduction}
\label{intro}
The deployment of machine learning (ML) models in sensitive domains---including criminal justice, healthcare, advertising, and finance--- is increasingly raising potential legal and fairness concerns \cite{mishraky2022bias}. In a number of studies involving sensitive information, bias was detected in ML models that were used to make decisions. For example, racial discrimination was detected in the COMPAS recidivism assessment model \cite{angwin2016machine} and Google's Ads model \cite{sweeney2013discrimination}; gender discrimination was found in Amazon's recruiting model \cite{dastin2018amazon}; and skin tone bias was found in models used to detect melanoma in images \cite{adamson2018machine}. Because the results produced by ML models rely on the data on which they are trained, any pre-existing societal biases could be embedded in data and may be inherited by these models unless proper mitigants are applied. There is a growing body of research focused on combating unfairness in ML models from which a number of definitions of fairness have been advanced, and many tools have been developed to satisfy them. Two of the most prominent categories of fairness definitions include: (1) group fairness \cite{hardt2016equality, chouldechova2017fair,chen2020towards,feldman2015certifying,kamiran2012data,calders2010three,kamishima2012fairness} and (2) individual fairness \cite{dwork2012fairness, kusner2017counterfactual,joseph2016fairness,friedler2016possibility,jung2019eliciting, ilvento2019metric, mukherjee2020two}. The output of an ML model is a distribution of scores, between 0 and 1, to which a threshold is applied to produce decisions. Group fairness definitions focus on ensuring that different demographic groups are treated equally, meaning that these decisions are unbiased based on group affiliation. Individual fairness definitions aim to guarantee that individuals with similar feature values are treated similarly, meaning that an ML model assigns them similar scores. 

As multiple works have shown~\cite{chouldechova2017fair, kleinberg2016inherent, zhao2022inherent}, achieving a universal notion of fairness is impossible. Compromises must be made when it is desirable to satisfy multiple conflicting notions of fairness. How to balance these trade-offs depends on the particular application. For example, unless a discrete exception applies, it is prohibited under U.S. fair lending laws\footnote{The Equal Credit Opportunity Act and its implementing regulation, Regulation B, prohibit creditors from considering any protected characteristic (such as race, gender, or age, etc.) in any aspect of a credit transaction, unless an express exception applies.} to take into consideration in any aspect of a credit transaction protected attributes, including, but not limited to, race, sex, and religion\cite{RN124}. Thus, financial institutions undertake qualitative and quantitative fair lending analyses to identify, assess, and mitigate any associated fair lending risks \cite{chen2019fairness}. The European Union also prohibits discrimination on the basis of a similar set of protected attributes under Title III (Equality), Article 21 of the European Union Charter of Fundamental Rights~\cite{kilpatrick2022article}. 

A variety of approaches have been proposed to preserve group-based fairness~\cite{Hsu,chen2020towards,chen2020transparency, zafar2017fairnessA, zafar2017fairnessB, hardt2016equality, jiang2020wasserstein}. To offset biases reflected in the score distributions produced by a model for different demographic groups, Chen et al. construct an optimization problem to solve for separate scoring thresholds for each group to produce fair decisions~\cite{chen2020transparency}. Hardt et al.~\cite{hardt2016equality} and Hsu et al.~\cite{Hsu} approach this problem by equalizing the output scores of different groups by constructing post-processing transformations that are applied to the original distribution of scores produced by an ML model. However, these approaches assume that sensitive group affiliation information is directly available to their decision-making models for processing in the testing stage of the ML process, which is often not the case and, under some circumstances, may be prohibited by applicable law. That is to say, given the restrictive, highly sensitive nature of protected attributes, it is common to restrict the model development team's access to those attributes. Instead, that team develops the models that will be integrated into the institution’s decision-making process, while a separate compliance team with access to the sensitive data evaluates the models for fairness. As such, the models do not make direct use of the sensitive attributes, but those attributes are available to the compliance team that tests the models for fairness.

Regularization-based methods provide a potential solution to this bottleneck~\cite{chen2020towards,chen2020transparency, zafar2017fairnessA, zafar2017fairnessB, jiang2020wasserstein}. Regularization is a common method used in many AI/ML applications~\cite{tian2022comprehensive,xiong2022grod, zhao122017marginalized} to improve model generalization or teach models to satisfy additional constraints. In fairness-aware AI/ML applications, regularization is often used to teach a model to enforce group fairness by decorrelating its decisions with respect to the sensitive attribute during the training process. Yet these methods do not directly include the sensitive attribute as an input to the model, meaning that it is only used in the testing phase of the ML process, not the training phase. However, a lack of heterogeneity in the feature values of individuals in different demographic groups may impede a model from learning how to equalize different demographic groups' score distributions without disparately impacting similar individuals within each group. Hence, asking a classifier to approximate statistical parity between different groups may cause it to badly violate parity among the individuals within a group~\cite{kearns2018preventing}.

This highlights an important consideration that must be addressed in dealing with problems in which group fairness is a hard constraint---solutions that satisfy group fairness must treat individuals \textit{within} each group fairly with respect to each other. Two issues must be addressed to satisfy this requirement: (1) a definition must be constructed to articulate what it means for individuals within the same group to be treated fairly (within-group fairness) and (2) a solution must be developed to satisfy within-group fairness without violating group fairness. This paper addresses these two issues, particularly under the constraining scenario in which the sensitive attribute is not allowed to be directly provided to a decision-making model in the testing phase of the ML process. In the remainder of this paper, we will use the terminology \textit{inter-group fairness} in place of group fairness to avoid any potential confusion between this notion and the notion of \textit{within-group fairness}.

To achieve our two aforementioned goals, we take advantage of a key insight: though an ML model is prohibited from directly using a sensitive attribute in the testing phase of the ML process, it may be available to \textit{pre-process the data} prior to providing it to the ML model for decision-making. Referring to our example on fair lending, this means that a compliance team could pre-process the data it oversees before providing it to the machine learning team for use. Thus, in this paper, we introduce a pre-processing framework for simultaneously satisfying inter- and within-group fairness. The core idea of our pre-processing framework is to devise a mapping that (1) removes the correlation between different groups' feature distributions while (2) ensuring that the ordering of scores produced by an ML model for individuals within the same group is preserved for the pre-processed feature distribution. We adopt a stringent notion of fairness---the threshold invariant form of demographic parity \cite{kamiran2012data}---for our inter-group fairness definition \cite{chen2020towards} and introduce a new definition of \textit{within-group fairness}, which is used to devise a measure for fairness assessment. Furthermore, we compare this approach against regularization-based training methods to verify its utility. To summarize, this paper provides the following contributions:
\begin{itemize}
    \item We introduce a new definition of fairness, termed \textit{within-group fairness}, which ensures that individuals within the same group are treated fairly with respect to each other.
    \item We provide a metric for measuring within-group fairness.
    \item We introduce a novel framework that uses pre-processing to achieve both inter- and within-group fairness.
    \item We experimentally verify that our pre-processing framework can achieve both inter- and within-group fairness with little compromise in accuracy, while empirically demonstrating that regularization methods struggle to satisfy both without compromising model performance.
\end{itemize}

The remainder of this paper is organized as follows. In Section \ref{background}, a summary of basic fairness-related concepts relevant to this work is provided. In Section \ref{prob_setup}, the problem setup is formulated, while in Section \ref{methods}, the pre-processing framework is described along with the regularized training approach used for comparison. Experimental results are presented in Section \ref{results} to quantify each method's accuracy and ability to achieve inter-group fairness and within-group fairness. A discussion of the limitations associated with this paper is provided in Section~\ref{limits_mits}. Finally, the paper is concluded in Section \ref{conclusion}.

\section{Background and Related Work}
\label{background}

\subsection{Fairness definitions} 
Fairness definitions may be grouped into three main categories, which include inter-group, individual, and subgroup fairness~\cite{mehrabi2021survey}. Inter-group fairness is measured by prediction performance parity across different demographic groups. To deal with multifaceted issues of bias and discrimination, various inter-group fairness notions have been proposed in the literature, including demographic parity~\cite{dwork2012fairness}, equalized odds~\cite{hardt2016equality}, test fairness\cite{chouldechova2017fair}, and threshold invariant fairness~\cite{chen2020towards}. However, solutions that only enforce inter-group fairness may produce outcomes that are blatantly unfair from the perspective of an individual~\cite{zemel2013learning}. Individual fairness aims to ensure that similar prediction performance is achieved for similar individuals with respect to the same task~\cite{dwork2012fairness}. To bridge the gap between inter-group and individual fairness, Kearns et al.~\cite{kearns2018preventing} propose the notion of subgroup fairness. This definition enforces inter-group fairness on a large collection of subgroups defined over combinations of protected attribute values. 

In this paper, we define a new notion of fairness in Section~\ref{withinfair}, termed \textit{within-group} fairness, which aims to preserve the scoring hierarchy of individuals from the same demographic group before and after accounting for inter-group fairness. This idea is related to the sub-group fairness definition proposed by Kearns et al. \cite{kearns2018preventing} in the special case where each individual in a group is considered a sub-group.

\subsection{Design of fairness-aware algorithms}
Many methods have been studied to achieve some notion of fairness in machine learning, which can be categorized accordingly: (1) pre-processing, (2) in-processing, and (3) post-processing~\cite{mehrabi2021survey}. Pre-processing approaches remove the underlying biases in the raw feature data prior to providing it to an ML model for training or decision-making. Kamiran et al.~\cite{kamiran2012data} propose a pre-processing method that ``massages” a set of training labels to remove bias in it prior to training a model. In their method, a ranker is used to select which labels to alter while minimizing deterioration in prediction accuracy. In-processing methods introduce fairness-aware regularizers into the objective function during the training process, which aim to strike a balance between maximizing accuracy and minimizing unfairness~\cite{kamishima2012fairness}. Calders et al.~\cite{calders2010three} devised a regularizer that requires a trained classifier to make decisions independent of the sensitive attribute, while Zafar et al.~\cite{zafar2017fairnessA,zafar2017fairnessB} enforce demographic parity and equalized odds constraints in the classifier training process. Chen et al.~\cite{chen2020towards} designed a fairness loss function that aims to equalize the score distributions of different demographic groups. Post-processing achieves fairness by reassigning the scores produced by the initial black-box model by applying a function to the original output scores~\cite{feldman2015certifying}. Other post-processing methods~\cite{mehrabi2021attributing} can mitigate bias by identifying unfair features via attention mechanisms and manipulating the corresponding attention weights.

In this paper, we propose a novel pre-processing framework that maps the feature vectors of different demographic groups to a \textit{canonical feature distribution} in which inter- and within-group fairness can simultaneously be achieved with low cost to a model's performance. For a comprehensive review of algorithmic fairness that provides a discussion of the prominent causes of unfairness, algorithms, definitions, and datasets available in this field of study, we refer the reader to Pessach et al.~\cite{pessach2020algorithmic}.

\section{Problem Setup}
\label{prob_setup}

In this section, we outline the problem that we aim to solve. Our end goal is to devise an algorithm that can achieve both inter- and within-group fairness without compromising predictiveness. To achieve this goal, we now introduce the inter- and within-group fairness notions that we want our model to satisfy and the measures, $\Delta_{TIDP}$ and $\Delta_{WGF}$, that quantify how well a method is able to satisfy inter- and within-group fairness, respectively.   

\subsection{Notation}

Following Dwork et al., we refer to \textit{individuals} as the objects that we aim to classify \cite{dwork2012fairness}. We refer to the set of individuals as $\mathcal{I}$. Each individual, $i \in \mathcal{I}$, has a corresponding \textit{feature vector}, $\mathbf{f}_i \in \mathbb{R}^{k}$, the elements of which represent values associated with features from the \textit{feature set}, $\mathcal{F} \triangleq \{f_1, \hdots, f_k\}$. We refer to a subset of features, $\mathcal{A}\subseteq\mathcal{F}$ as \textit{sensitive} if we are prohibited from discriminating against individuals based on the values of these features. Thus, no sensitive feature values are included in feature vectors. Each individual also has a  \textit{label}, $y\in\{0,1\}$, associated with him or her. A \textit{dataset} of $N$ individuals, $X \in \mathbb{R}^{N \times k}$, is a matrix in which each row represents an individual's feature vector. The primary task of binary classification is to predict the labels of individuals using a scoring function. A scoring function, $s_{\boldsymbol{\theta}}:\mathbb{R}^k\to[0,1]$, is defined as a mapping of a feature vector to a score between $0$ and $1$, inclusive. Given a threshold, $t$, the estimator associated with a scoring function, $s_{\boldsymbol{\theta}}$, is represented by the function:

\begin{equation}
    \hat{Y} = \begin{cases}
    0,& s_{\boldsymbol{\theta}}(\mathbf{f})\leq t\\
    1,& s_{\boldsymbol{\theta}}(\mathbf{f})> t,
\end{cases}
\end{equation}
and is used to predict the label associated with a given individual. A scoring function is trained on data to optimize its parameters to satisfy an objective and tested on separate data to ensure the results it produces are generalizable. We refer to the training dataset as $\mathbf{X}_{tr}$ and the testing dataset as $\mathbf{X}_{ts}$. A \textit{loss function}, $L_{\boldsymbol{\theta}}$, is used to quantify the ability of the scoring function to produce scores that satisfy the objective. A \textit{learning algorithm} optimizes the weights of the scoring function, $\boldsymbol{\theta}$, according to $L_{\boldsymbol{\theta}}$.

A \textit{group}, $g\subseteq{\mathcal{I}}$, is a collection of individuals that share the same values for the sensitive features in $\mathcal{A}$. The set of all groups is denoted $\mathcal{G}$. We use numerical superscripts to specify group membership. For example, $X^{(i)}$ refers to the subset of the dataset $X$ belonging to group $i$. The omission of superscripts indicates that we are referring to population information. Lastly, $d_{\mathcal{X}}$ represents a distance measure over the set $\mathcal{X}$.

\subsection{Inter-group Fairness}
\label{interfair}
A variety of definitions have been proposed to capture inter-group fairness. Two of the most prominent definitions include demographic parity (DP) and equalized odds (EO), and we focus on DP in this paper.

\begin{definition}
    (Demographic Parity) An estimator, $\hat{Y}$, satisfies demographic parity for a binary feature, $A\in\{0,1\}$, if
    $P(\hat{Y}=1|A=0) = P(\hat{Y}=1|A=1)$.
\end{definition}

An important observation to make about this definition is that it is threshold dependent. That is, a particular classifier may achieve DP for a given threshold, but not for all thresholds in the range $[0,1]$ when the group score distributions produced by a scoring function deviate from each other. In many applications, 
this may lead to fairness issues. As described in Section I, for example, model development teams may not have access to sensitive demographic information associated with the data. Without such information, they may produce models with unequal output score distributions across groups, raising potential fairness concerns. Thus, we wish to ensure that a model produces score distributions that are equal across groups without consideration of any protected attributes. This intuition is encapsulated in the more strict definition of \textit{threshold-invariant fairness} defined by Chen et al. \cite{chen2020towards} (also coined \textit{strong demographic parity} by Jiang et al. ~\cite{jiang2020wasserstein}), and has been employed in a variety of works to enforce inter-group-fair classification~\cite{chen2020towards,jiang2020wasserstein,chen2020transparency,silvia2020general,ramaswamy2021fair}  and regression~\cite{silvia2020general,chzhen2020fair}. Thus, we use this definition for the inter-group fairness notion in our analysis:

\begin{definition}
    (Threshold Invariant Fairness)  Threshold
Invariant Demographic Parity (TIDP) is achieved when DP is satisfied, independent of the decision threshold t.
\end{definition}

To measure how well a classifier preserves TIDP, we can take the average of the Calder-Verwer (CV) score \cite{calders2010three} for every possible threshold value $t\in[0,1]$, where the CV scores for DP are given by:
$|P(\hat{Y}=1|A=0) - P(\hat{Y}=1|A=1)|$. We use $\Delta_{TIDP}$ to refer to this resulting average value.
The framework that we propose in this paper aims to achieve TIDP while maintaining within-group fairness.

\subsection{Within-group Fairness}
\label{withinfair}

In the previous subsection, we discussed measures of inter-group fairness, which are blind to the treatment of individuals. To understand the negative side-effects that may arise from only taking inter-group fairness into account, we provide the following simple motivating example.

\begin{figure}
  \begin{center}
    \includegraphics[width=0.75\textwidth]{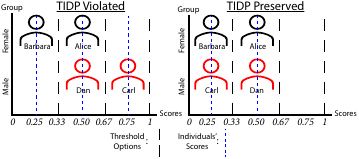}
  \end{center}
\caption{Illustration of Fairness in Example \ref{ex1}.}
\label{ex1_fig}
\end{figure}

\begin{example}
\label{ex1}
Consider four individuals, Alice and Barbara, who are female, and Carl and Dan, who are male. Assume that all other sensitive attributes are the same for all individuals. According to the information in their loan applications, they receive the following loan approval scores, respectively: 0.50, 0.25; 0.75, 0.50. Since the distribution of scores between the male and female groups is different, a loan officer adjusts Carl's score to a value of 0.25 so that TIDP-based inter-group fairness is satisfied since for any threshold, the same number of individuals from each sex score above it.
(see Figure~\ref{ex1_fig}). However, assuming all applicants were scored according only to non-sensitive features, we see that Carl's outcome is unfair since he has a stronger application than Dan.
\end{example}

This simple example highlights an important consideration that must be taken into account in the fairness process: individual fairness must still be preserved within each group. Clearly, a more individually fair way to achieve TIDP in Example \ref{ex1} would be to provide Carl with a score of 0.50 and Dan with a score of 0.25. While the specific example we provide here deals with finance, such considerations could potentially impact recidivism assessment~\cite{larson2016we}, hiring decisions~\cite{Langenkamp}, credit approvals~\cite{intahchomphoo2020artificial}, and any other real-world classification applications in which bias has been detected in the score distributions produced by an ML model. Thus, we introduce a new fairness definition, \textit{within-group fairness}, which takes advantage of the insight provided in this example: the most accurate scores are the fairest for individuals that belong to the same group. This suggests that the scores provided by a \textit{baseline} model, which does not explicitly account for any notion of fairness, yield the fairest scores when comparing individuals within the same group. 

The inspiration for our definition comes from Dwork et al.'s definition of individual fairness, which states that similar individuals should be treated similarly \cite{dwork2012fairness}. Definition~\ref{ind_fair}, below, provides the formal definition of individual fairness.
\begin{definition}
\label{ind_fair}
(Individual Fairness) A scoring function $s:\mathbb{R}^k\to[0,1]$ satisfies individual fairness if for any two individuals $i,j \in \mathcal{I}$,
$|P(\hat{Y}_{i} = y)-P(\hat{Y}_{j} = y)|\leq \epsilon; \ \ \ \ \ if   \ \ d_{\mathcal{I}}(i,j) \approx 0$.
\end{definition}

With this in mind, we provide a few definitions that build toward our definition of within-group fairness.

\begin{definition} (Signed Distance Function)
    The signed distance function for $x, y \in \mathcal{X}$ is given by:\\
    \begin{equation}
        \phi_{\mathcal{X}}(x,y) =
        \begin{cases}
            d_{\mathcal{X}}(x,y),& x \leq y\\
            -d_{\mathcal{X}}(x,y),& x > y
        \end{cases}
    \end{equation}
\end{definition} 

\begin{definition} (Individual Fairness Across Mappings)
    A mapping $Q:\mathbb{R}^k\to[0,1]$ satisfies individual fairness with respect to a different mapping $R:\mathbb{R}^k\to[0,1]$ if for any two individuals, $i,j \in \mathcal{I}$,
    \begin{equation}
        |\phi_{[0,1]}(Q(\mathbf{f}_{i}) , Q(\mathbf{f}_{j})) - \phi_{[0,1]}(R(\mathbf{f}_{i}) , R(\mathbf{f}_{j}))| \leq \epsilon.
    \end{equation}
\end{definition}

\begin{definition}\label{WGF} (Within-group Fairness Across Mappings)
    A mapping $Q:\mathbb{R}^k\to[0,1]$ is said to satisfy within-group fairness with respect to a different mapping $R:\mathbb{R}^k\to[0,1]$ if for any group $g \in \mathcal{G}$, individual fairness across models is satisfied between any two individuals, $i,j \in g$. That is,
    \begin{align}
        |\phi_{[0,1]}(Q(\mathbf{f}_{i}) , Q(\mathbf{f}_{j})) - \phi_{[0,1]}(R(\mathbf{f}_{i}) ,& R(\mathbf{f}_{j}))|  \leq \epsilon, \notag\\
        &\forall g\in \mathcal{G}; \forall i,j\in g
    \end{align}
\end{definition}

Intuitively, our within-group fairness definition states that the scores provided by one mapping are fair with respect to another if the distance between any two individuals' scores is relatively preserved and the ordering of their scores does not change across models. In the algorithm proposed in Section~\ref{methods}, we consider one of the mappings to be the baseline scoring function. The second mapping is considered to be a pre-processing transformation $T: \mathbb{R}^k\to\mathbb{R}^k$ composed with the baseline scoring function. Since we design $T$ to remove the bias between the distribution of feature vectors for different demographic groups, if Definition \ref{WGF} is satisfied, then our framework achieves both inter- and within-group fairness. To quantify how well a mapping, $Q$, is able to preserve within-group fairness with respect to another mapping, $R$, we use the following equation:
{\small
\begin{align}
    \label{wgf_metric}
    &\Delta_{WGF} =\notag\\ 
    &\frac{2\sum\limits_{g\in \mathcal{G}} \sum\limits_{\substack{i,j\in g\\i<j}} \mathbf{1}\{|\phi_{[0,1]}(Q(\mathbf{f}_i), Q(\mathbf{f}_j)) - \phi_{[0,1]}(R(\mathbf{f}_i), R(\mathbf{f}_j))| > \epsilon\}}{\sum\limits_{g\in \mathcal{G}}N_g(N_g-1)},
\end{align}}\\
\noindent where $\mathbf{1}$ represents the indicator function, $\epsilon$ may be a user-specified parameter quantifying how much within-group unfairness we are willing to tolerate, and $N_g$ is the number of individuals in group $g$.
$\Delta_{WGF}$ accounts for the change in signed distance between every pair of scores within each demographic group that results from using mapping $R$ in place of mapping $Q$. If the signed distance between the scores of any pair of individuals produced by mapping $Q$ is drastically different from the baseline model (i.e. greater than $\epsilon$), this means these individuals are being treated drastically differently. That is, within-group fairness is violated for this pair of individuals. Hence, the numerator of Equation~\ref{wgf_metric} counts all the pairs of individuals for which an inter-group-fair model violates within-group fairness, while the denominator counts how many pairs of individuals exist in each group. Thus, $\Delta_{WGF}$ provides the percentage of individuals for which within-group fairness is violated.

\section{Framework for Achieving Group and Within-group Fairness}
\label{methods}

In this section, we outline the proposed pre-processing framework for achieving inter- and within-group fairness. We also describe an in-processing algorithm in which fairness is encouraged through the loss function during training with the use of regularizers, which we use to compare against our framework.

\subsection{Proposed Pre-processing Framework}
\label{pre-process}

\begin{figure*}[t]
\centerline{\includegraphics[width=\columnwidth]{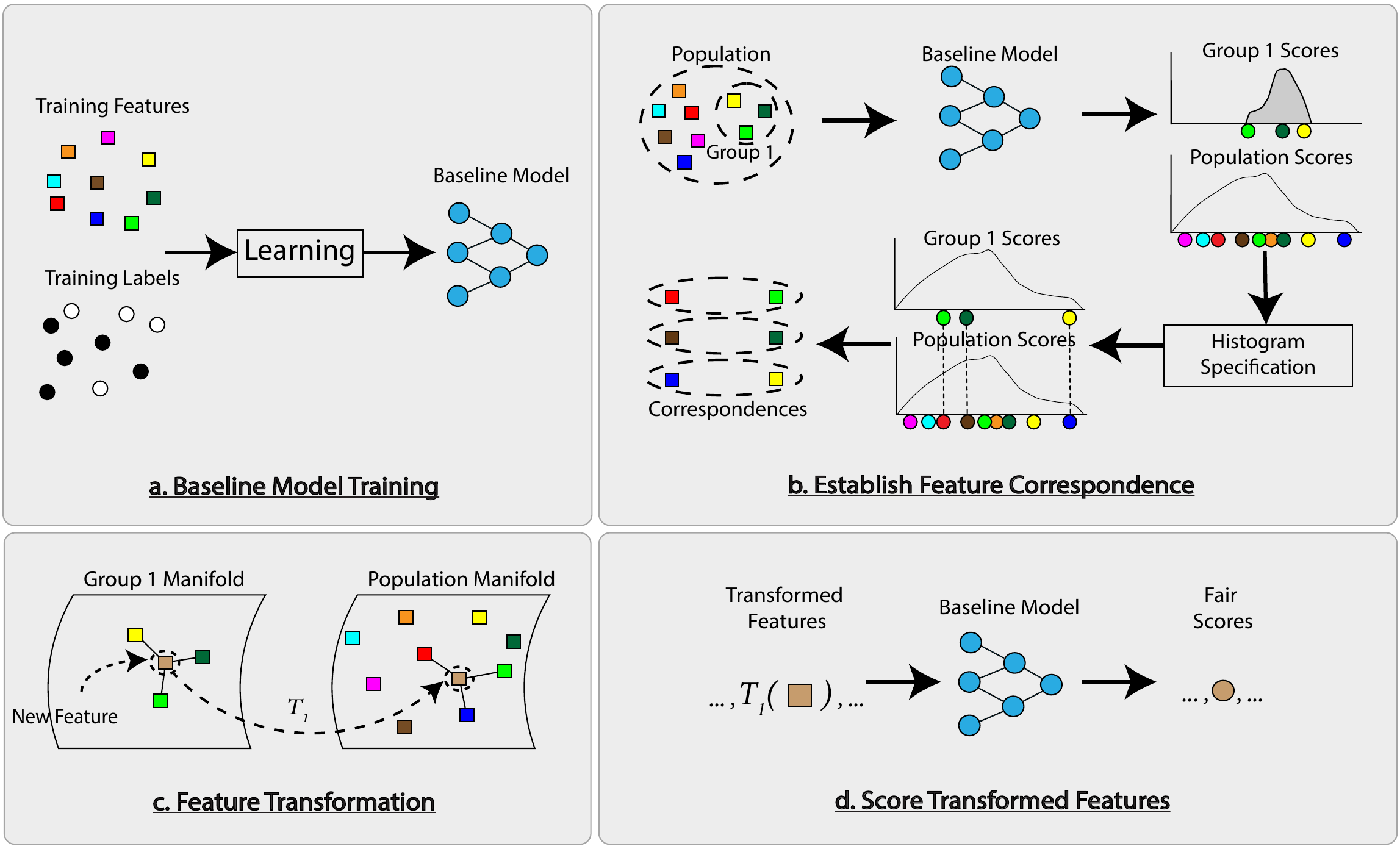}}
\caption{Overview of the pre-processing framework for achieving inter- and within-group fairness.}
\label{fig::algo_diagram}
\end{figure*}

Figure \ref{fig::algo_diagram} illustrates the proposed pre-processing algorithm. 
In Phase I, we train an ML scoring function on training data. This scoring function may produce disparate score distributions for different groups, violating inter-group fairness. To deal with this issue, we construct transformations that map the feature vectors of individuals from different groups to a domain in which the bias between these groups is removed, which we refer to as the \textit{canonical domain}. 
The goal is to design these mappings such that when the baseline scoring function is applied to the transformed feature vectors of any group, the output score distribution resembles the original un-transformed \textit{population} score distribution. Each group's transformation is constructed by creating a \textit{correspondence} between the feature vectors within the group and the feature vectors in the entire population. The details of each phase of the system are presented in the ensuing subsections.

\noindent \textit{\underline{Phase I: Baseline Model Training}}:
Assume we have a population training dataset, $\mathbf{X}_{tr}\in \mathbb{R}^{N_{tr} \times k}$ with associated labels $\mathbf{y}\in \{0,1\}^{N_{tr}}$ consisting of $N_{tr}$ individuals. We construct a scoring function of the form $s_{\boldsymbol{\theta}}(x) = \sigma(m_{\boldsymbol{\theta}}(x))$, where $m_{\boldsymbol{\theta}}:\mathbb{R}^k \to \mathbb{R}$ is an ML model (e.g. logistic regression, support vector machine, neural network, etc.) and $\sigma:\mathbb{R}\to[0,1]$ is the sigmoid function. We call this scoring function the \textit{baseline model}. The baseline model is designed to maximize \textit{accuracy} according to the training data using a standard (sub-) differentiable loss function, $L_{\boldsymbol{\theta}}$, (e.g., cross-entropy, hinge, etc.). Finally, we use a learning algorithm (e.g., gradient descent, stochastic gradient descent, etc.) to optimize the model weights, $\boldsymbol{\theta}$.

\noindent \textit{\underline{Phase II: Establish Feature Correspondence}}:
Once the baseline model, $s_{\boldsymbol{\theta}}$, is obtained from Phase I, we split the population training data into groups using the sensitive attribute, where we assume these groups are mutually exclusive. Specifically, we decompose $\mathbf{X}_{tr}$ into group training datasets $\mathbf{X}_{tr}^{(i)}, i=1,\hdots,|\mathcal{G}|$. The baseline model is applied to each group's training dataset and the population training dataset to obtain the score vectors $\mathbf{s}_{tr}^{(i)}, i=1,\hdots,|\mathcal{G}|$ and $\mathbf{s}_{tr}$, respectively. We then map the distribution of scores associated with each group to the population score distribution. To do this, we apply histogram specification~\cite{GonzalezW08}, which is a distribution matching technique commonly applied in image processing to transform the histogram of the pixel values in a given image to match a specified histogram. In our case, the histograms of the group and population score distributions are computed and histogram matching is applied to each group's distribution to match it to the population distribution. We denote these score vectors by $\hat{\mathbf{s}}_{tr}^{(i)}, i=1,\hdots,|\mathcal{G}|$. Because the scores in $\hat{\mathbf{s}}_{tr}^{(i)}$ and $\mathbf{s}_{tr}$ are approximately equally distributed, each score in $\hat{\mathbf{s}}_{tr}^{(i)}$ should be close in distance to some score in $\mathbf{s}_{tr}$. We pair group and population feature vectors if their scores are sufficiently close, calling such pairs \textit{feature correspondences}. Below, we provide the formal definition of a feature correspondence.

\begin{definition} (Feature Correspondence) 
    Let $a\in \hat{\mathbf{s}}_{tr}^{(i)}$ and $b \in \mathbf{s}_{tr}$ be the scores associated with the $n^{th}$ feature vector of group $i$, $\mathbf{f}_n^{(i)}$, and the $l^{th}$ feature vector of the population, $\mathbf{f}_l$, respectively. We say $\mathbf{f}_n^{(i)}$ and $\mathbf{f}_l$ form a correspondence, denoted $\mathbf{f}_n^{(i)} \sim \mathbf{f}_{l}$, if for any $c \in \mathbf{s}_{tr}$, such that $c\neq b$, the following inequality holds
    $|a-b| \leq |a-c|$.
    % \begin{equation}
    %     |a-b| \leq |a-c|.
    % \end{equation}
\end{definition}

Our goal is to exploit these feature correspondences to create a transformation that maps feature vectors from each group to the canonical population domain. This idea is illustrated in Figure \ref{fig::algo_diagram}b. Critically, histogram specification is a monotonically increasing transform on the scores. As a result, when we use these correspondences to construct our group transformations, it should follow that the scores that the baseline model produces for the transformed features preserve the ordering of the scores that the baseline model produced for the $original$ feature vectors. As a result, our notion of within-group fairness should be preserved.

\noindent \textit{\underline{Phase III: Mapping Construction}}:
Phase III uses the feature correspondences found in Phase II to construct a mapping between each group domain and the population domain for individuals not seen in the training data. Our approach assumes that each group's data lie on a manifold in $\mathbb{R}^k$ and that the training data sample this manifold densely enough to provide a faithful representation of it. Hence our goal is to construct a more general correspondence between each group's manifold to the population manifold. To achieve this goal, we note that the field of computational geometry provides a variety of candidate data structures that may be used for manifold representation to construct these mappings. Two notable candidates include: (1) the Delaunay Triangulation (DT), which serves as the dual counterpart of the Voronoi Diagram (VD), and (2) the $k$-$d$ tree.

\begin{figure}[t]
  \begin{center}
    \includegraphics[width=0.4\textwidth]{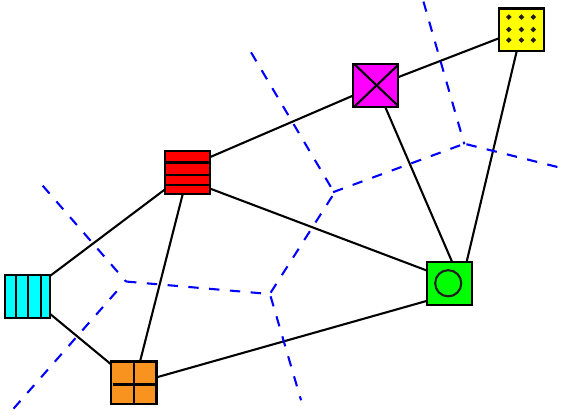}
  \end{center}
  \caption{Dual relationship between Voronoi diagram and Delaunay Triangulation.}
  \label{Vor}
\end{figure}

Given a set of points in $\mathbb{R}^k$ called \textit{cell cites}---which are given by the training feature vectors in our context, a VD induces a cell decomposition of space where each cell represents all of the points closest to a particular cell site. A DT is a planar subdivision of $\mathbb{R}^k$, constructed by drawing edges connecting any two cell sites that share an edge between their VD cells, and as a result, the closest pair of sites in a VD is represented as neighbors in the DT. In this way, each cell in the VD corresponds to a vertex in the dual complex. Figure~\ref{Vor} illustrates this primal-dual relationship. 

\begin{figure}[t]
  \begin{center}
    \includegraphics[width=0.65\textwidth]{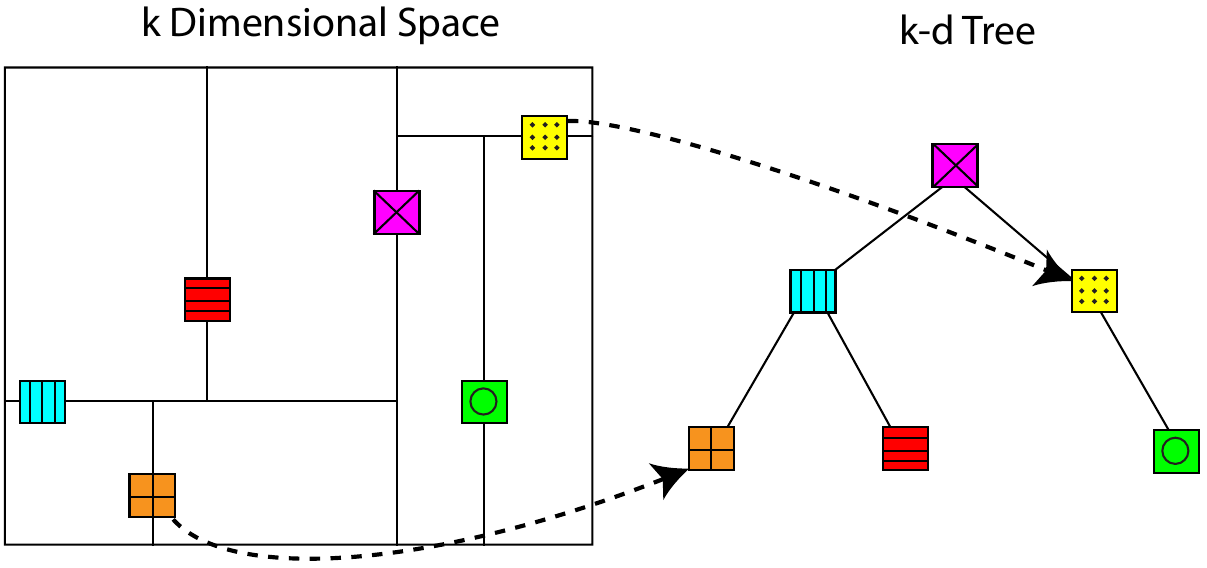}
  \end{center}
  \caption{Representation of a $k$-$d$ tree.}
  \label{kd_tree}
\end{figure}

The $k$-$d$ tree is a space-partitioning data structure used to organize points in $k$-dimensional space. In this data structure, each leaf node represents one data point, while each non-leaf node represents a splitting hyperplane used to generate two half-spaces. The points in each half-space correspond to the leaves of a given node’s subtree. This structure is illustrated in Figure~\ref{kd_tree}.

The utility of each of these data structures lies in their ability to represent new feature vectors as a weighted combination of the closest neighboring training feature vectors in $\mathbb{R}^k$ used to construct each data structure. This can be done by constructing $|\mathcal{G}|$  DT or $k$-$d$ tree data structures using the feature vectors in the training sets $\mathbf{X}_{tr}^{(i)}, i=1,\hdots,|\mathcal{G}|$ to represent each group's individual manifold.  When a new feature vector for group $i$, $\mathbf{f}^{(i)}_{new}$, is to be pre-processed, we can represent it by a weighted average of feature vectors to which it is closest, as illustrated in Figure \ref{fig::algo_diagram}c, which are obtained by traversing the given data structure to find the closest neighbors to represent the new feature.
In particular, mapping new feature vectors from the group domain to the population domain can be done as follows. Without loss of generality (WLOG), let $\{\mathbf{f}_n^{(i)}\}_{n=1}^{k}$ represent the $k$ nearest neighbors of $\mathbf{f}^{(i)}_{new}$, belonging to group $i$. The Euclidean distances between $\mathbf{f}^{(i)}_{new}$ and the elements of $\{\mathbf{f}_n^{(i)}\}_{n=1}^{k}$ are represented by $\{d_n^{(i)}\}_{n=1}^{k}$. Inspired by the algorithm presented by Dongsheng et al. \cite{an2019ae}, we  construct a set of weights $\{ {d_n^{(i)-1}}/{\sum_{l=1}^{k}d_l^{(i)-1}}\}_{n=1}^{k} = \{w_n^{(i)}\}_{n=1}^{k}$ 
associated with each element in $\{\mathbf{f}_n^{(i)}\}_{n=1}^{k}$ that reflect how closely they resemble the new feature vector in the group domain. In the population domain, WLOG, we obtain the corresponding feature vectors $\{\mathbf{f}_n\}_{n=1}^{k}$, where $\mathbf{f}^{(i)}_n \sim \mathbf{f}_{n}$, %for $n=1,\hdots,k$, 
and use these weights to construct their new feature vectors in the population domain $\mathbf{f}_{new} = \sum_{l=1}^{k}w_l f_{l}$. We use the notation $T^{(i)}:\mathbb{R}^{k}\to \mathbb{R}^{k}$ to describe the mapping from a feature vector from group $i$'s domain to the population domain. Hence, given a test dataset, $\mathbf{X}_{ts} \in \mathbb{R}^{N_{ts} \times k}$ of $N_{ts}$ individuals, we can decompose it in group test sets $\mathbf{X}_{ts}^{(i)}, i,...,|\mathcal{G}|$, obtain the canonical group test dataset, $T^{(i)}(\mathbf{X}_{ts}^{(i)}), i,...,|\mathcal{G}|$, and apply the baseline model, $s_{\boldsymbol{\theta}}$, to it. A user-specified threshold, $t$, can then be used to make the classification decision for these test data.

In our pre-processing framework, we utilize the $k$-$d$ tree instead of the DT because it possesses one important property: it can be used to \textit{efficiently} perform nearest neighbor searches for data in \textit{high dimensions}. Conversely, performing searches using DTs in higher dimensions can be extremely expensive. We elaborate on the significance of this property in Section~\ref{manifold_choice} and provide runtime analyses for each data structure to illustrate the importance of this property.

\begin{figure*}
    \includegraphics[width=0.98\textwidth]{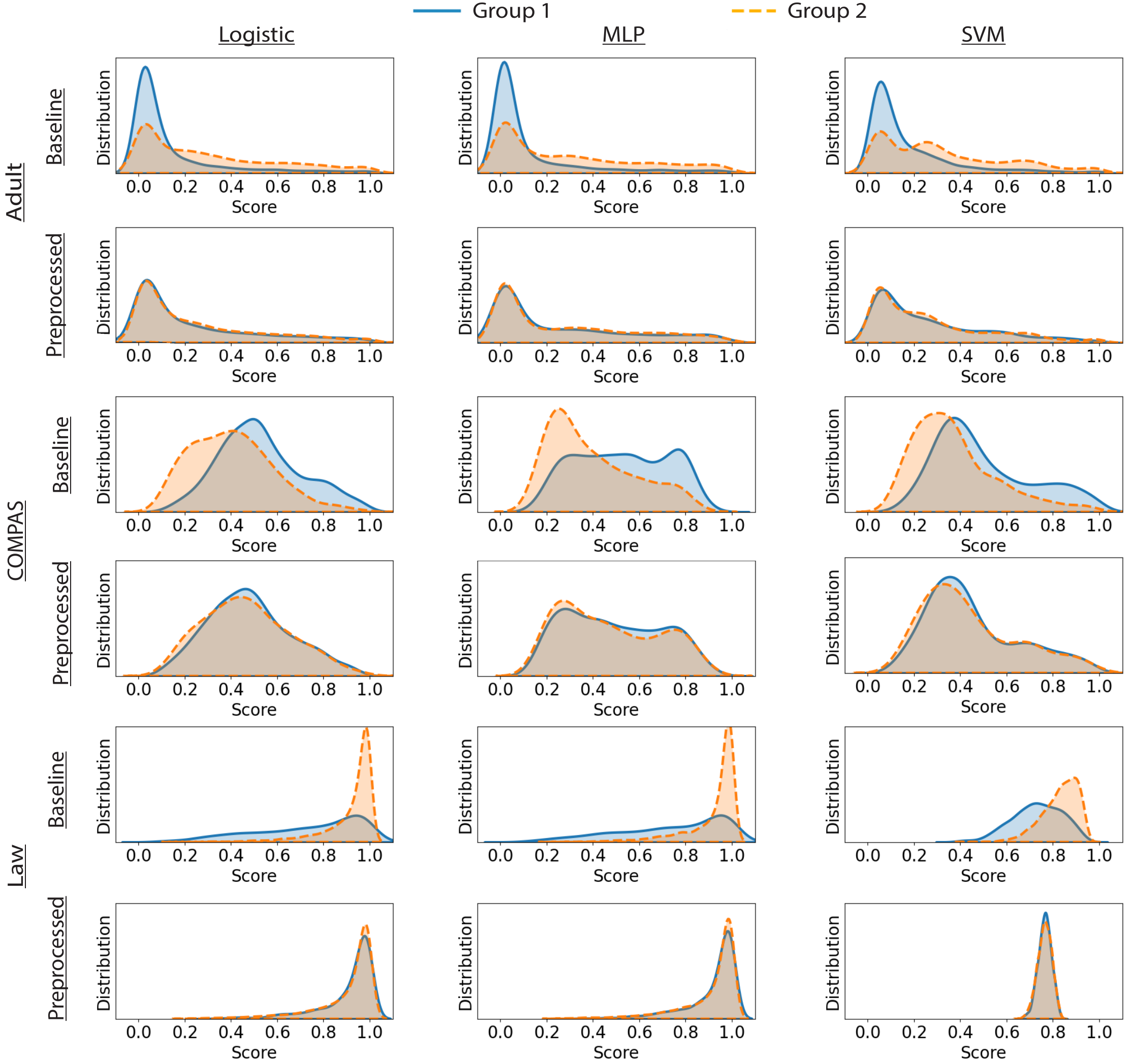}
    \caption{The distribution of scores produced for each dataset by the baseline scoring function and the pre-processing framework for using three different ML models.}%
    \label{dists}
\end{figure*}

\subsection{In-processing Framework}
\label{inprocess}

In this section, we describe an in-processing alternative to our pre-processing framework for comparison. In this approach, a scoring function is trained to internalize inter-group fairness in the training process. This approach is similar to that of the algorithm described by Chen et al.~\cite{chen2020towards}.

We start by training a baseline scoring function, $s_{\boldsymbol{\theta}}(x)$, as in Phase I of Section \ref{pre-process}. The weights $\boldsymbol{\theta}$ serve as a starting point from which we will \textit{fine-tune} to improve the fairness of the scoring function. Specifically, once we obtain
\begin{equation}
    \boldsymbol{\theta}^{base}=\underset{\boldsymbol{\theta}}{\arg\min}  \ L_{\boldsymbol{\theta}}(\mathbf{X}_{tr}),
\end{equation}
We learn another set of weights, $\boldsymbol{\theta}^{eq}$, for $s_{\boldsymbol{\theta}}(x)$ through
\begin{equation}
    \label{finetune}
    \boldsymbol{\theta}^{eq}=\underset{\theta}{\arg\min}  \ L_{\boldsymbol{\theta}}(\mathbf{X}_{tr}) + \lambda E(\mathbf{X}_{tr}),
\end{equation}
by initializing  $\boldsymbol{\theta} = \boldsymbol{\theta}^{base}$ before applying the learning algorithm. Here, $E(\mathbf{X}_{tr})$, is a regularizer that uses a distance notion to calculate the distance between the scores from two different groups, $i,j \in \mathcal{G}$. We take this two-stage fine-tuning approach instead of directly optimizing Equation \ref{finetune} starting from random weights for two reasons. First, by using $\boldsymbol{\theta}^{base}$ as our starting point, we are able to start at the most accurate solution and observe the trade-off we incur between inter-group fairness and accuracy during the training process in \ref{finetune}. Second, training the scoring model using Equation \ref{finetune} with a randomized initialization of $\boldsymbol{\theta}$ can lead to instability in the resulting output of $\boldsymbol{\theta}^{eq}$ reflected by the random seed from which $\boldsymbol{\theta}$ is initialized.

Two common regularizers are considered to compare with our pre-processing framework: (1) an Earth Mover's distance (EMD) regularizer~\cite{hou2016squared} and (2) a  Kullback–Leibler (KL) divergence regularizer~\cite{chen2020towards}. The EMD regularizer is of the form: 
\begin{equation}
    EMD(\mathbf{X}_{tr}^{(i)},\mathbf{X}_{tr}^{(j)}) = \sum_{n=1}^{C}(CDF_n(\hat{\mathbf{h}}_i)-CDF_n(\hat{\mathbf{h}}_j))^2,
\end{equation}
where
$\hat{\mathbf{h}}_i$ and $\hat{\mathbf{h}}_j$ are Gaussian approximated histogram bins calculated from $\mathbf{X}_{tr}^{(i)}$ and $\mathbf{X}_{tr}^{(j)}$, $CDF_{n}$ represents the energy in the $n^{th}$ bin of the cumulative distribution function, and $n$ represents the bin number. Because rectangular histogram bins are non-differentiable at the bin edges, we use Gaussian approximations of the rectangular histogram bins as done by Chen et al.~\cite{chen2020towards} so that we can apply a (sub-)gradient-based learning algorithm for optimizing $\boldsymbol{\theta}$.

For the second regularizer tested, we use the symmetric KL divergence with an added Gaussian assumption that Chen et al.~\cite{chen2020towards} formulate. Since the KL divergence is less tractable than the EMD, using the aforementioned Gaussian approximated histogram binning approach leads to convergence issues when trying to equalize group distributions. Adding the Gaussian assumption instead leads to a simple loss that is only dependent on the second-order statistics of each group dataset, alleviating these tractability issues. Thus, letting $\mathcal{N}_i$ and $\mathcal{N}_j$ represent normal distributions with the same mean and variance as group $i$ and group $j$, respectively. Then, the regularizer is given as:
\begin{align}
KL&(\mathbf{X}_{tr}^{(i)},\mathbf{X}_{tr}^{(j)}) \notag\\
&=  
\frac{1}{2}( \frac{(\hat{\mu}_i-\hat{\mu}_j)-\hat{\sigma}_i}{\hat{\sigma}_j} + \frac{(\hat{\mu}_j-\hat{\mu}_i)-\hat{\sigma}_j}{\hat{\sigma}_i})-2.
\end{align}

\section{Experimental Results}
\label{results}

\begin{figure*}
    \includegraphics[width=0.95\textwidth]{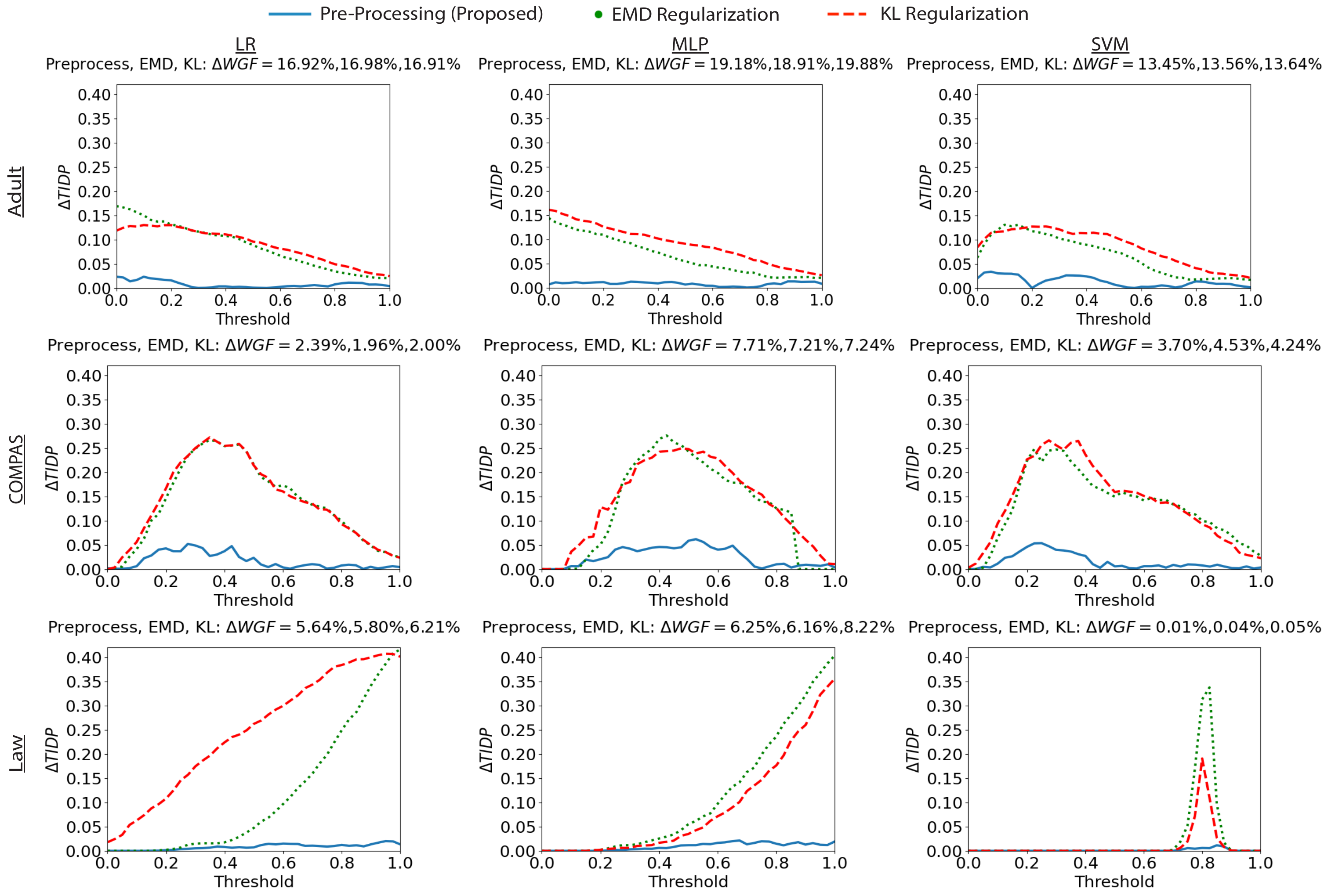}
    \caption{Plots of the value of $\Delta_{TIDP}$ at different thresholds when an average $\Delta_{WGF}$ value has been achieved for three different models: (a) LR, (b)MLP, and (c) SVM.
}%
\label{TIDPplt}
\end{figure*}

\begin{table}[t]
\caption{Dataset Compositions}
\label{DTA_comp}
\centering
\resizebox{0.65\columnwidth}{!}{%
\begin{tabular}{ c ||c|c| c}
 \hline
 Characteristic& Adult & COMPAS& Law\\
 \hline
 Features & \begin{tabular}{@{}c@{}} age \\ workclass \\ fnlwgt \\ education \\ education-num \\ marital-status \\ occupation \\ relationship \\ race \\  capital-gain \\  capital-loss \\ hours-per-week \\ native-country \end{tabular} & \begin{tabular}{@{}c@{}} age \\ sex \\ juv\_fel\_count \\ juv\_misd\_count \\ juv\_other\_count \\ priors\_count \\ c\_charge\_degree\end{tabular}  & \begin{tabular}{@{}c@{}} decile1b \\ decile3 \\ lsat \\ ugpa \\ zfygpa \\ zgpa \\ fulltime \\ family income \\ sex \\ tier \end{tabular} \\ 
 \hline
 Class Label & \begin{tabular}{@{}c@{}}Income >50k \\ (0/1)\end{tabular} &\begin{tabular}{@{}c@{}}Recividate \\ (0/1)\end{tabular} & \begin{tabular}{@{}c@{}}Pass Bar \\ (0/1)\end{tabular} \\
 \hline
 Sensitive attribute & \begin{tabular}{@{}c@{}}Sex \\ (Male/Female)\end{tabular} & \begin{tabular}{@{}c@{}}Race \\ (White/Black)\end{tabular}  &  \begin{tabular}{@{}c@{}}Race \\ (White/Non-white)\end{tabular} \\
 \hline
 \begin{tabular}{@{}c@{}}\# Samples \\ (Group 1/Group 2)\end{tabular} & 30527 / 14695 & 2103 / 3175 & 17491 / 3307\\
 \hline
 \begin{tabular}{@{}c@{}}\# 0/1 Class Labels \\ (Group 1)\end{tabular} & 20988 / 9539 & 822 / 1281 & 1377 / 16114\\
 \hline
 \begin{tabular}{@{}c@{}}\# 0/1 Class Labels \\ (Group 2)\end{tabular} & 13026 / 1669 & 1514 / 1661 & 916 / 2391\\
 \hline
\end{tabular}
}
\end{table}
\subsection{Datasets for Experimental Studies}
In this section, the proposed pre-processing framework is analyzed on the Adult~\cite{kohavi1996scaling}, COMPAS risk assessment~\cite{larson2016we} and Law School~\cite{wightman1998lsac} datasets---three benchmark datasets with known biases with respect to a given sensitive attribute. For brevity, we simply refer to these datasets as the Adult, COMPAS, and Law datasets. The Adult dataset contains a variety of features correlated with an individual's financial status. The ML task for this dataset is to predict whether individuals in the dataset make above or below \$50,000 annually, with bias existing in the features on the basis of sex, with females being less likely than males to make over \$50,000. The COMPAS dataset is composed of the records of criminal defendants screened by the COMPAS model in Broward County. It includes information on a defendant's demographics, case details, and criminal histories, with the ML task being to use this information to predict whether a person recidivated within two years of being screened. Biases have been found to exist in the score distributions of white and black demographic groups produced by ML models trained on this dataset, with the white demographic group having lower scores than the black demographic group on average. The Law dataset contains law school admissions records from a survey conducted by the Law School Admission Council (LSAC) across 163 law schools in the United States in 1991. Using the variables provided in this dataset, the ML task is to predict whether someone is likely to pass the bar exam or not. ML models trained on this dataset have been shown to produce biased score distributions when predicting whether white and non-white demographic groups will pass the bar examination, with the white demographic group having much higher scores on average. 

In Table~\ref{DTA_comp} we provide details on the composition of each dataset used for training and testing the models for our experiments. Full descriptions of the features in each dataset are provided in Table ~\ref{all_desc}
%, \ref{COMPAS_desc}, and \ref{Law_desc}
of the appendix. Group 1 and Group 2 respectively refer to the advantaged and disadvantaged groups of each dataset. It can clearly be observed that bias exists in the class label distributions associated with each group. We also note that 0 and 1 class labels associated with each dataset may have opposing connotations. For the Adult and Law datasets, a 1 class label respectively indicates that a person makes over \$50,000 annually or passed the bar examination (which is good), while for the COMPAS dataset, a 1 class label indicates that a person did recidivate (which is bad). Moreover, in the COMPAS dataset, the minority group (Group 1) is advantaged, while the opposite is the case in the Adult and Law dataset. We can further observe that there is a large variation in the sample size of each dataset. Thus, there is diversity in the distributions associated with each dataset, which is good for drawing generalized insights from our results.

\subsection{Experimental Details}
The ensuing subsections are devoted to testing our pre-processing framework's ability to achieve both inter- and within-group fairness on all three datasets. We compare its performance with the regularization approaches described in Section~\ref{inprocess}. As done by Chen et al.~\cite{chen2020towards}, we split all data into 70\% training and 30\% testing for the COMPAS and Law datasets and make the simplifying assumption that race is the only binary sensitive attribute that creates inter-group biases in the data. The Adult dataset was originally partitioned into a 75\%-25\% training-testing split by its creators. For this dataset, we treat sex as the sensitive attribute. All binary and categorical variables were one-hot-encoded and all ML models were trained in TensorFlow using gradient descent with step sizes between 0.1 and 0.0001. All programs were conducted using Python 3.8 and run on a MacBook Pro (1.7 GHz Quad-Core Intel Core i7) with no GPU support. We used the SciPy library for $k$-$d$ tree construction and set $k=10$ for the number of nearest neighbors used in constructing the mappings from the group to population domain for all experiments. We also use this library to construct Delaunay triangulations for the experimental analysis provided in Section~\ref{manifold_choice}.

\subsection{Baseline vs. Pre-processing Framework: Risk Distribution Comparison}

In this section, the risk score distributions produced by the baseline scoring function and the pre-processing framework are visually compared for three different scoring functions: a logistic regression (LR), three-layer perception (MLP), and support vector machine (SVM). These distributions are presented in Figure \ref{dists}, with the first and second rows corresponding with the Adult dataset, the third and fourth rows corresponding with the COMPAS dataset, and the fifth and sixth rows corresponding with the Law dataset. The first, third, and fifth rows of distributions capture the baseline scoring function results applied to the raw feature vectors. The second, fourth, and sixth rows of distributions use the pre-processing framework to map the raw feature vectors to the canonical population domains for each dataset before applying their baseline scoring functions to generate their score distributions. The dark, overlapped regions in these plots capture the similarity between the distributions of Groups 1 and 2. A classifier that perfectly captures the threshold invariant demographic parity group fairness definition, TIDP, would produce entirely overlapped distributions. It can clearly be seen that distributions associated with the pre-processed features show significantly more overlap than those associated with the raw features for each model for all datasets. Notably, the biases seen in the risk distributions for all sensitive groups is removed when the canonical features are used in place of the raw features for each dataset. Hence, these illustrations capture the improvements made in inter-group fairness by applying the baseline scoring function to the canonical features instead of to raw features. 

\begin{figure*}
    \includegraphics[width=0.95\textwidth]{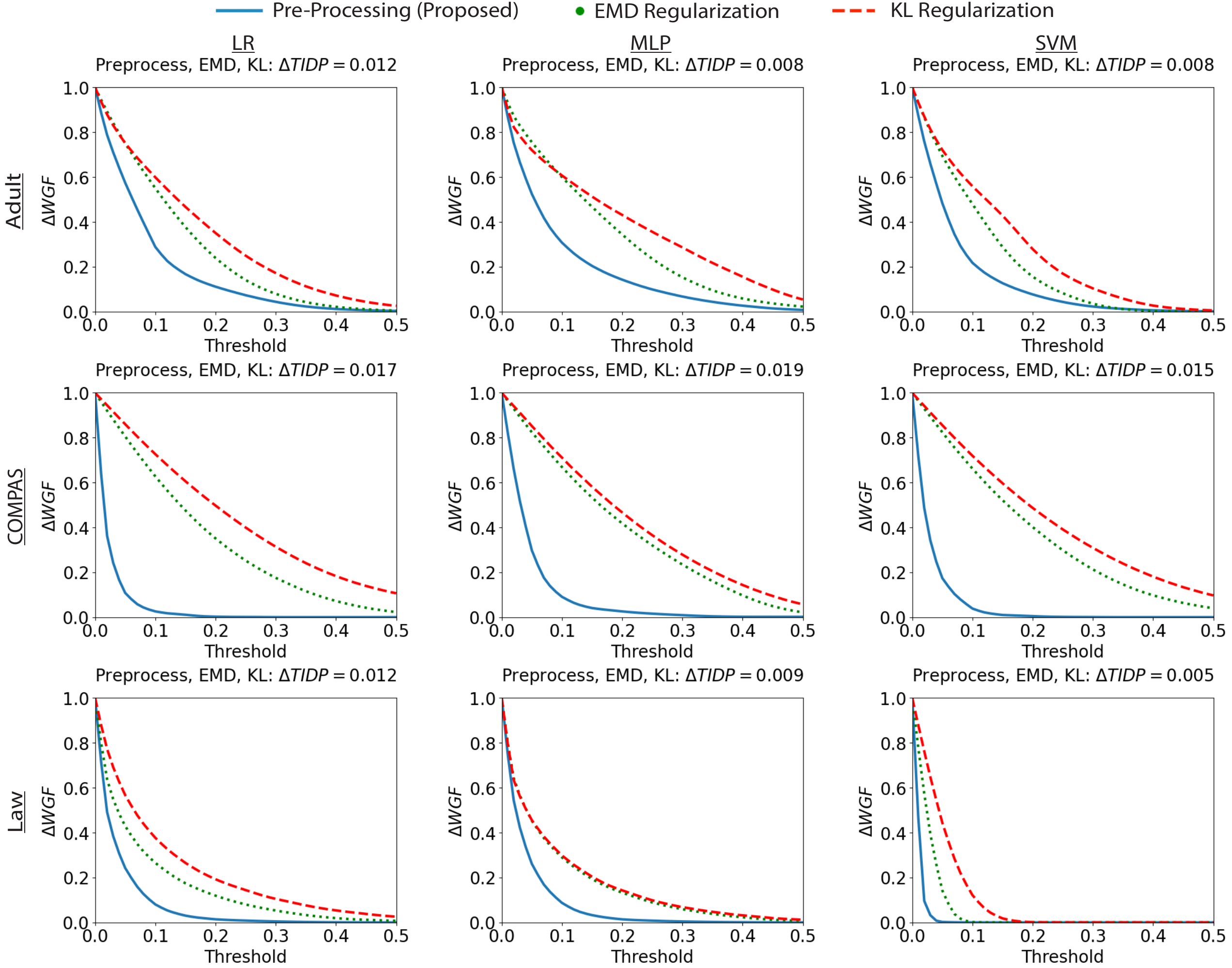}
    \caption{Plots of $\Delta_{WGF}$ at different thresholds when an average $\Delta_{TIDP}$ value has been achieved for three different models: (a) LR, (b)MLP, and (c) SVM.
}%
\label{WGFplt}
\end{figure*}

\subsection{Pre-processing Framework vs. Regularization Methods}
\label{results_pre_vs_reg}
In this section, our pre-processing framework's ability to achieve inter- and within-group fairness is compared with the two regularization methods described in Section \ref{inprocess} respectively using the measures $\Delta_{TIDP}$ and $\Delta_{WGF}$ from Sections~\ref{interfair} and ~\ref{withinfair}. $\Delta_{TIDP}$ represents the average Calder-Verwer (CV) score obtained for all thresholds $t\in[0,1]$. The CV score measures the absolute difference between the percentage of individuals in each demographic group that lie on one side of the threshold decision boundary, $t$. The lower this value, the better a model is at satisfying inter-group fairness for a given value of $t$. By averaging over all values of $t\in[0,1]$, $\Delta_{TIDP}$ ensures that inter-group fairness is not sensitive to an arbitrary choice of $t$. Hence smaller values of $\Delta_{TIDP}$ indicate that a model is able to satisfy inter-group fairness regardless of our choice in $t$, meaning that the distributions of scores for different demographic groups are equalized.

As described in Section~\ref{withinfair}, $\Delta_{WGF}$ measures the percentage of individuals that are treated differently by two different mappings (i.e. the change in the signed distance between pairs of scores within a group is greater than $\epsilon$). This means that a lower value of this metric leads to better preservation of the treatment of individuals across two mappings. In our case, we require any proposed method to treat pairs of individuals within the same demographic group similarly to the baseline model to ensure that within-group fairness is preserved. 

Intuitively, obtaining low values of $\Delta_{TIDP}$ may require a model to adjust two groups' score distributions to overlap. In situations in which a model applies large morphological changes to each group’s score distribution (meaning the ordering of scores or shapes of the group distributions is significantly changed), this will cause tension between  $\Delta_{TIDP}$ and $\Delta_{WGF}$. This is because $\Delta_{WGF}$ aims to preserve the structure of each group’s baseline score distribution. Thus, a model that can successfully achieve low values of $\Delta_{TIDP}$ and $\Delta_{WGF}$ will minimize the structural changes required to equalize each group’s score distributions.

Regularization methods achieve different levels of $\Delta_{TIDP}$ and $\Delta_{WGF}$ through tuning the hyper-parameter, $\lambda$, in Equation~\ref{finetune}. In contrast, our pre-processing framework achieves a single level of fairness since it involves no hyper-parameters. Therefore, we aim to compare the ability of these approaches to simultaneously satisfy inter- and within-group fairness by answering the following questions: 
(1) How do the values of $\Delta_{TIDP}$ compare when the regularization methods are required to achieve comparable values of $\Delta_{WGF}$ to our pre-processing method? (2) How do the values of $\Delta_{WGF}$ compare when the regularization methods are required to achieve comparable values of $\Delta_{TIDP}$ to our pre-processing method?
If the regularization methods are able to perform as well as our pre-processing framework, the answer to question~(1) should be that the differences between groups, measured by $\Delta_{TIDP}$, across all models are comparable, and the answer to question~(2) should be that the mapping differences, measured by $\Delta_{WGF}$, across all models are comparable.

We approach the first question as follows. For each pair of models and regularizers, we perform multiple rounds of training over a range of values for $\lambda$ in Equation~\ref{finetune}. Then, from all the training sessions for each pair, we select the model for which the average value (taken over $\epsilon\in [0, 0.5]$) of $\Delta_{WGF}$ is closest to equaling the average value of $\Delta_{WGF}$ 
produced by our pre-processing method. Given that the pre-processing framework and regularization methods achieve approximately equal values of $\Delta_{WGF}$, the method that produces a lower value of $\Delta_{TIDP}$ is better at achieving overall fairness. The second question is approached in the reverse direction. That is, we again perform multiple rounds of training for each model-regularizer pairing. Then, from all the training sessions for each pair, we select the model for which the average value (taken over $t\in [0, 1]$) of $\Delta_{TIDP}$ is closest to the average value of $\Delta_{TIDP}$ 
produced by our pre-processing method. We refer the reader to Section~\ref{sens_analysis} for a detailed analysis of $\lambda$.

The results of our comparative analysis are summarized in Figure~\ref{TIDPplt} and Figure~\ref{WGFplt} for the Adult, COMPAS, and Law testing data. Figure~\ref{TIDPplt} displays plots of the inter-group fairness measure, $\Delta_{TIDP}$, for the pre-processing framework and both regularization methods for three different ML models. The values of $\Delta_{WGF}$ above each plot represent the average value of the within-group fairness measure associated with a particular method. Thus, we can see that for approximately equal values of $\Delta_{WGF}$, the pre-processing framework consistently provides significantly lower $\Delta_{TIDP}$ values for all models and thresholds for all datasets.

Figure~\ref{WGFplt} displays plots of the within-group fairness measure, $\Delta_{WGF}$, for the pre-processing framework and both regularization methods for three different ML models. The value of $\Delta_{TIDP}$ above each plot represents the average value of the inter-group fairness measure associated with a particular method. Again, regardless of the ML model, our pre-processing framework can achieve lower values $\Delta_{WGF}$ for a wide range of thresholds for all datasets, indicating that fewer pairs of individuals in the pre-processing framework violate within-group fairness. Thus, the results in Figures~\ref{TIDPplt} and \ref{WGFplt} indicate that the regularization methods must make significantly larger compromises between inter-group and within-group fairness than our pre-processing framework. 

\subsection{Performance Evaluation of Pre-processing Framework}
\label{Perform_eval}
In this section, we compare the performances of our pre-processing framework to the regularization methods when all methods achieve comparable levels of inter-group fairness (i.e. $\Delta_{TIDP}$ is approximately the same for each approach). The baseline classifier should produce superior performance to all of these methods. However, the performance of a fair model should not significantly deteriorate if it is to be considered useful in practice. Thus, we use two metrics---accuracy and area under the curve (AUC) for the receiver operating characteristic (ROC) curve---to compare the performances of our pre-processing framework and regularization methods to the baseline classifier's performance. 

To analyze the accuracy of a method, we use a value of 0.5 as the threshold for all classifiers since this is the most common value used in practice. The resulting performances of these methods on the test data of the Adult, COMPAS, and Law datasets are shown in Table~\ref{accuracy_all}. To verify that each approach achieves a comparable level of inter-group fairness, we provide the values of $\Delta_{TIDP}$ associated with each model in this table. The results show that the pre-processing framework outperforms the regularization-based methods in AUC for all datasets and accuracy for all datasets except the Adult dataset, for which the EMD regularizer produces slightly higher accuracy. Compared with the baseline model, no more than a 2.78\% drop in accuracy and 0.022 drop in AUC is incurred by any of the three ML models when applying the pre-processing framework to the COMPAS dataset. In contrast, when applying the regularization methods to the COMPAS dataset, we see accuracy drop-offs of over 8.91\% and AUC drop-offs of over 0.100 in all cases. 

Although the pre-processing framework also outperforms the regularization approaches on the Law dataset, the difference is less pronounced. That is, all methods incur a little drop in accuracy from the baseline model. This is because there is large class imbalance for this dataset, with 88.86\% of the class labels in the test set and 88.97\% of the class labels in the entire dataset having a value of 1. Thus, simply assigning everyone the same class label will achieve a high accuracy. This explains why the accuracies of all regularization methods for the Law dataset are identical---each of the regularized models assigns a class label of 1 to every person in the test set. On the other hand, the accuracies displayed for our pre-processing method for the Law dataset show slightly higher variability. The higher accuracies produced by the LR and MLP models result from our pre-processing method's ability to preserve the left tails of the score distributions produced by their baseline models. This is illustrated in the first and second plots in the bottom row of Figure~\ref{dists} in which the tails of the score distributions produced by the pre-processing method stretch below the 0.5 threshold. As a result, some of the test samples are assigned 0 class labels. Conversely, the AUC for the ROC curve shows much more significant drop-offs in the results obtained for the regularization methods than the pre-processing method. This is because the ordering of scores produced by the regularization methods has become shuffled with respect to the score ordering of the baseline model. This causes the results to be sensitive to the placement of the decision threshold over the support of the score distribution.

As previously mentioned, our pre-processing approach achieves slightly lower accuracy on the Adult dataset compared to the EMD regularizer when using a 0.5 classification threshold. This is not unexpected since accuracy is determined by applying a single threshold to the score distributions to perform classification. Though 0.5 is a common choice of threshold, it may not be universally optimal for every dataset. This observation is reflected in the fact that our pre-processing framework's AUC performance is higher than the EMD regularizer's for every dataset, including the Adult dataset. Since AUC captures the dynamics associated with selecting any threshold in the range $[0,1]$, it is threshold invariant.

\begin{table}
\caption{Performance comparisons on the Adult, COMPAS, and Law Datasets. Results are presented across five trials.}
\label{accuracy_all}
\includegraphics[width=0.95\textwidth]{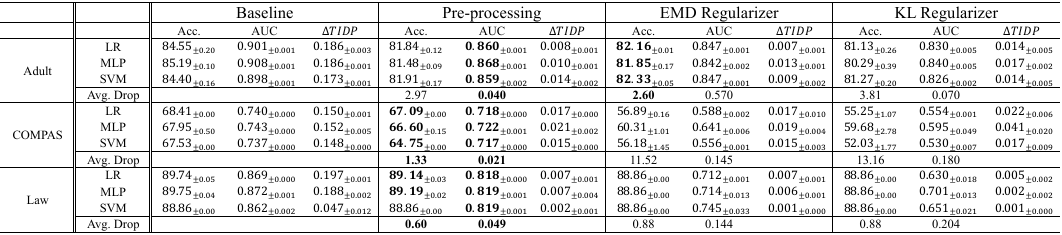}
\end{table}

\subsection{$\lambda$ Sensitivity Analysis}
\label{sens_analysis}

\begin{figure*}
    \includegraphics[width=0.95\textwidth]{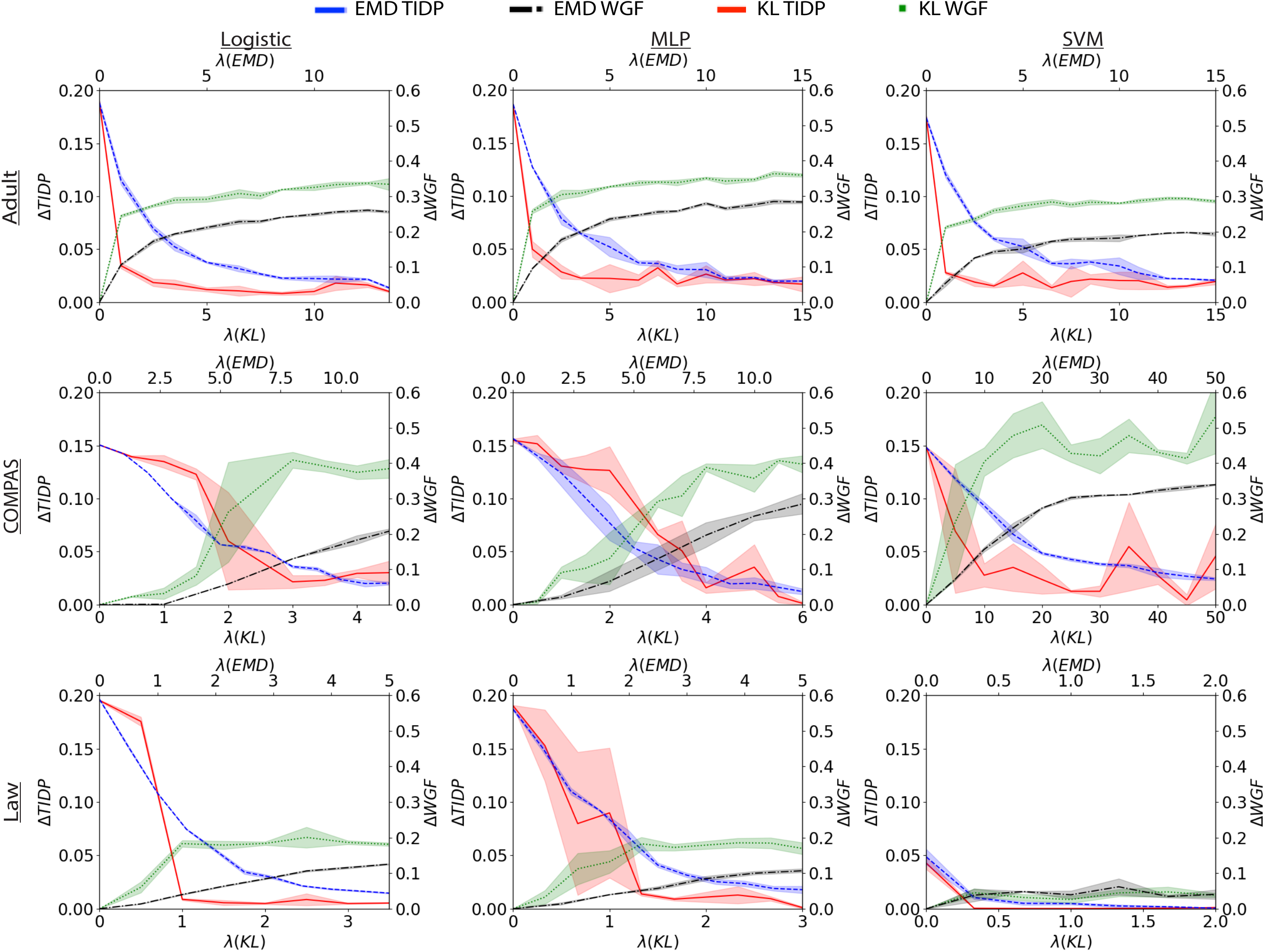}
    \caption{Plots of $\Delta_{TIDP}$ and $\Delta_{WGF}$ for the EMD and KL regularizers for different values of $\lambda$. Results provided for training LR, MLP, and SVM models over 5 sessions for the Adult, Compas, and Law datasets.
}%
\label{sens_plots}
\end{figure*}

In Section~\ref{results_pre_vs_reg}, we analyzed  inter- and within-group fairness for the EMD and KL regularization methods by tuning $\lambda$ in Equation~\ref{finetune}. In this section, we provide experimental results that illustrate the effect that $\lambda$ has on the values of $\Delta_{TIDP}$ and $\Delta_{WGF}$ for these methods. Figure~\ref{sens_plots} provides plots that illustrate the progression of the values of $\Delta_{TIDP}$ and $\Delta_{WGF}$ as $\lambda$ is tuned over a range of values to train the LR, MLP, and SVM models. For all datasets, each of these models was trained over five sessions for each value of $\lambda$. Each plot contains four curves (with standard deviation error bars from the five trials) which provide the test results for the values of $\Delta_{TIDP}$ and $\Delta_{WGF}$ for the EMD and KL regularizers. The left and right y-axes respectively provide the scales for the $\Delta_{TIDP}$ (blue and red) and $\Delta_{WGF}$ (black and green) curves. We personalize the tuning of $\lambda$ for each model and regularizer pairing since the strength of $\lambda$ may have different effects on each pairing. Thus, the top x-axis provides the scale of $\lambda$ for all EMD regularizer results (the blue and black curves), while the bottom x-axis provides the scale of $\lambda$ for all KL regularizer results (the red and green curves). 

As the strength of $\lambda$ increases, we can see that the red and blue curves decrease in value, meaning that the values of $\Delta_{TIDP}$  associated with the two regularizers drop, and group fairness is improved. However, in every case, drops in $\Delta_{TIDP}$ lead to increases in $\Delta_{WGF}$, indicated by the green and black curves, which means that there is an increase in the violation of within-group fairness. The size of this effect varies by dataset and model. The COMPAS dataset, for example, tends to lead to the largest violation of $\Delta_{WGF}$ and also produces the least stable curves, indicated by the large standard deviations associated with its curves. The features in this dataset likely lack enough heterogeneity to allow a model to learn to distinguish between its two sensitive groups, causing it to make extreme adjustments to its baseline score distribution to satisfy inter-group fairness. These results highlight the contribution of our pre-processing framework, which does not suffer this inter- and within-group fairness trade-off. The Adult and Law datasets also suffer from this issue, though the extent of these violations is less pronounced. This is in part because these contain a larger and more fruitful set of features.

\subsection{Time Complexity Comparison: $k$-$d$ Tree vs Delaunay Triangulation}
\label{manifold_choice}
An important and non-trivial consideration that must be taken into account for representing the manifold used in our pre-processing framework is the time required to construct the data structure used to represent it. While we have found that using the Delaunay Triangulation (DT) in place of the $k$-$d$ tree in our pre-processing framework produces comparable performance results for the COMPAS dataset, the time complexity for computing a DT from $N$ feature vectors in dimension $k$ is $\mathcal{O}(N^{\lceil k \rceil})$, which quickly becomes intractable as the feature vector dimension grows to double digits. In comparison, the time complexity of constructing a $k$-$d$ tree is $\mathcal{O}(kNlog(N))$, which is far less expensive and linear in the feature vector dimension, $k$. To illustrate this computational difference, we present the time complexities in seconds associated with computing these data structures for Groups 1 and 2 for the Adult, COMPAS, and Law datasets in Table~\ref{time_complexity}. A clear and stark contrast between the costs associated with computing the DTs and $k$-$d$ trees can clearly be observed in this table. In particular, the $k$-$d$ tree is always able to be computed in under 20 seconds for each group for all datasets. Conversely, no DT is able to be constructed in a relatively similar amount of time. Notably, for the Adult and Law datasets, each group's associated DT is unable to be computed in under 12 hours. This is because both of these datasets contain far more samples than the COMPAS dataset. This emphasizes the scalability issues associated with using the DT for manifold representation in higher dimensions. It is also notable that it costs more time to construct the $k$-$d$ tree for the Adult dataset than for the Law dataset. The reasons for this are: (1) the Adult dataset contains more samples and (2) the Adult dataset contains more categorical features than the Law dataset, meaning that when they are one-hot encoded, the dimensions of the associated feature vectors increase. Nevertheless, the time complexities associated with constructing the $k$-$d$ tree for the Adult dataset are still quite efficient.

\begin{table*}
\caption{Time complexity comparison for constructing data structures. Results are presented in seconds across five trials. Processes not terminated within 12 hours are reported as "---."}
\label{time_complexity}
\centering
\includegraphics[]{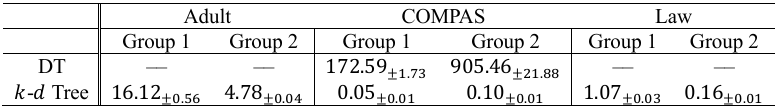}
\end{table*}

\section{Limitations and Discussions}
\label{limits_mits}
In this section, we list and discuss the limitations associated with our proposed framework. We first note that for the proposed algorithm to succeed in achieving both inter- and within-group fairness, the following assumption must approximately hold: if the distance between the scores of two feature vectors is close, then the distance between these feature vectors in the input space should be close. This assumption holds well on for the Adult, COMPAS, and Law datasets, but may not hold in general. If this situation fails, feature vectors that are close in the group domain may map to feature vectors that are far from each other in the population domain. Thus, and convex sum of these population feature vectors may lie outside the population manifold. To resolve this issue in our ongoing research, we are investigating new approaches for constructing feature correspondences between the group and population manifolds.

Second, we note that the mapping constructed in the configuration of the pre-processing framework proposed in this paper requires storing the entire training dataset in the $k$-$d$ tree data structure. While this approach allowed us to construct an interpretable mapping between a group and the population manifold, more compact representations may potentially be constructed by incorporating the mapping construction into the learning process. We plan to further investigate such solutions in our future work.

We also would like to observe that the focus of this work has been analyzing the effectiveness of our framework on tabular data. Nevertheless, our framework has the potential to be generalized to other forms of data, such as image data. While such data in its raw form may exist in much higher dimensions, which may produce a computational burden when constructing the $k$-$d$ tree, our framework may be applied to the features in the latent space of intermediate layers in an ML model or to feature vectors to which dimensionality reduction has first be applied.

Finally, we would like to mention two comments with regard to the results of our study. First, while we have observed that $TIDP$ and $WGF$ may conflict for the Adult, COMPAS, and Law datasets, this conflict is distributional dependent and should not be present in datasets for which the baseline classifier does to produce a major violation of demographic parity. For reference, Le Quy et al.~\cite{le2022survey} provide a survey of the primary benchmark datasets used for analyzing fairness in ML and list common fairness definitions that conflict with each. Second, we would like to acknowledge that, while we have compared our fairness framework against two regularization methods using three ML models, this list does not include an exhaustive analysis of all potential regularizers and ML models, which may influence our results.

\section{Conclusion}
\label{conclusion}

In this paper, we study the importance of maintaining within-group fairness in situations where satisfying inter-group fairness is required. We introduce a notion of within-group fairness and a metric for measuring it. We have adopted the stricter threshold invariant form of demographic parity for analyzing inter-group fairness, which requires the scores generated for different groups to be equally distributed. Furthermore, we introduce a novel pre-processing framework that maps raw features to a canonical domain before applying a classifier and performs well in achieving inter-group and within-group fairness. This framework requires sensitive group attributes to be used only for pre-processing the raw features in the testing stage of the machine learning process and never explicitly provides the sensitive attributes to a classifier to make decisions. To verify its effectiveness, we compare the performance of our framework to models in which fairness is embedded in the training process through regularization. Experimental results demonstrate that our pre-processing framework can achieve both inter-group and within-group fairness with little penalty on accuracy. 

\section{Acknowledgement}
We thank the reviewers for their insightful comments and feedback. This work was supported in part by the NSF grant \#IIS-2147276. We gratefully acknowledge support from the JP Morgan Chase AI Faculty Research Award.

\vspace{0.3cm}

\noindent \textit{Disclaimer:} This paper was prepared for informational purposes in part by the Artificial Intelligence Research group of JPMorgan Chase \& Co and its affiliates (``J.P. Morgan'') and is not a product of the Research Department of J.P. Morgan.  J.P. Morgan makes no representation and warranty whatsoever and disclaims all liability, for the completeness, accuracy or reliability of the information contained herein.  This document is not intended as investment research or investment advice, or a recommendation, offer or solicitation for the purchase or sale of any security, financial instrument, financial product or service, or to be used in any way for evaluating the merits of participating in any transaction, and shall not constitute a solicitation under any jurisdiction or to any person, if such solicitation under such jurisdiction or to such person would be unlawful.

\bibliographystyle{ieeetr}
\bibliography{refs}
\appendix

\section{Appendix}
In this appendix section, we provide details on the meaning of each of the features used for classification in the Adult~\cite{kohavi1996scaling}, COMPAS~\cite{larson2016we}, and Law~\cite{wightman1998lsac} datasets. These descriptions are respectively provided in Table~\ref{all_desc}.

\begin{table}[ht]
\caption{Feature descriptions for the Adult, COMPAS, and Law datasets}
\label{all_desc}
\centering
\includegraphics[width=0.65\linewidth]{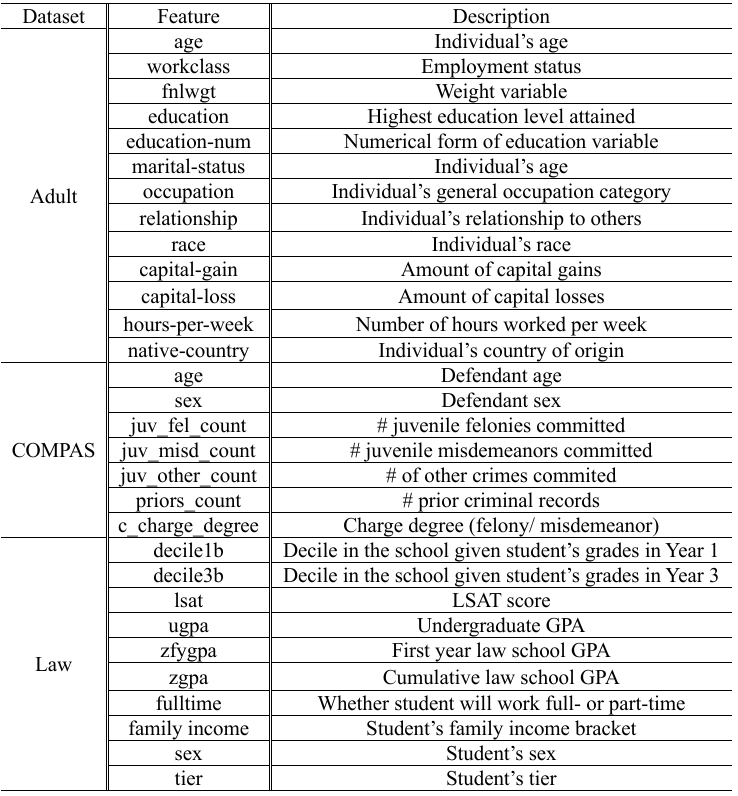}
\end{table}

\end{document}